\documentclass{amia}

\usepackage[labelfont=bf]{caption}
\usepackage[square]{natbib}
\setcitestyle{authoryear, square, semicolon}
\usepackage{color}
\usepackage{bbm}
\usepackage[]{authblk}
\usepackage[symbol]{footmisc}

\usepackage{graphicx}
\usepackage{subfigure}
\usepackage{booktabs}
\usepackage[table]{xcolor}

\usepackage{amsmath}
\usepackage{amssymb}
\usepackage{amsthm}
\usepackage{float}
\usepackage{braket}
\usepackage{graphicx}
\usepackage{hyperref}
\usepackage[ruled,vlined]{algorithm2e}

\usepackage[nameinlink,noabbrev,capitalise]{cleveref}    
\usepackage[retainorgcmds]{IEEEtrantools}

\newtheorem{theorem}{Theorem}
\newtheorem{proposition}{Proposition}
\newtheorem{assumption}{Assumption}
\crefname{assumption}{Assumption}{Assumptions}

\newcommand{\calD}{\mathcal{D}}
\newcommand{\calG}{\mathcal{G}}

\newcommand{\calQ}{\mathcal{Q}}

\newcommand{\Cat}{\mathrm{Categorical}}
\newcommand{\VI}{\mathrm{VI}}

\DeclareMathOperator*{\argmin}{arg\:min}
\DeclareMathOperator{\smsum}{\Sigma}

\def \pf {\theta}
\def \prho {w}
\def \bF {\mathbf{F}}

\def \x {\mathbf{x}}

\begin{document}

\title{Preferential Mixture-of-Experts:  Interpretable Models that Rely on Human Expertise As Much As Possible}

\author{Melanie F. Pradier, PhD$^{1,2,}$\footnote[1]{Equal contribution.}, Javier Zazo, PhD$^{1,2,\ast}$, Sonali Parbhoo, PhD$^{1,\ast}$,\\ Roy H. Perlis, MD MSc$^{3,4}$, Maurizio Zazzi, MD$^{5}$, Finale Doshi-Velez, PhD$^{1}$}
\institutes{
    $^1$School of Engineering and Applied Sciences, Harvard University, Cambridge, MA, USA;
    $^2$Health Intelligence, Microsoft Research, Cambridge, Cambridgeshire, UK; \\
    $^3$Center for Quantitative Health, Massachusetts General Hospital, Boston, MA, USA;
    $^4$Harvard Medical School, Boston, MA, USA
    $^5$University of Sienna, Italy.\\
}

\maketitle

\noindent{\bf Abstract}
\textit{We propose Preferential MoE, a novel human-ML mixture-of-experts model that augments human expertise in decision making with a data-based classifier only when necessary for predictive performance. Our model exhibits an interpretable gating function that provides information on when human rules should be followed or avoided.
The gating function is maximized for using human-based rules, and classification errors are minimized. We propose solving a coupled multi-objective problem with convex subproblems. We develop approximate algorithms and study their performance and convergence. Finally, we demonstrate the utility of Preferential MoE on two clinical applications for the treatment of Human Immunodeficiency Virus (HIV) and management of Major Depressive Disorder (MDD).}

\section{Introduction}
In the last few years, there has been a growth in the use of machine learning (ML) methods for decision-making in complex domains such as loan approvals, medical diagnosis and criminal justice. In particular, ML currently plays a key role in the healthcare sector for several tasks such as developing medical procedures \citep{hamid2017machine,esteva2017dermatologist}, handling patient data and records \citep{pianykh2020improving} and treating chronic diseases \citep{raghu2019algorithmic}.  However, these algorithms typically require large amounts of data to make reasonable predictions.  Additionally in the health sector, variability in practice between clinicians, patient heterogeneity, 
different disease prevalences, and confidentiality issues all result in final training cohorts being relatively small. 
Moreover, a clinician is often faced with rare events or outlier cases, where classic ML approaches suffer from insufficient training samples. 
In each of these scenarios, it is crucial to be able to incorporate clinical experience and domain knowledge.  

Specifically, in practice, clinicians often rely on relatively simple human-based rules that reflect reasonable approaches to handle a situation.
These rules can be seen as an additional source of knowledge that can be leveraged when building ML systems for clinical decision-support.
For instance, clinicians treating patients with HIV tend to adhere to a list of guidelines for administering first and second-line therapies specified by several organizations~\citep{WHO19,AAHIV}; other well-known guidelines exist for prescribing antidepressants to address Major Depressive Disorder (MDD)~\citep{lage2018heuristics}. Often, these rules provide benefits that are not easily formalized into a machine learning objective, for example, in terms of safety~\citep{stone_risk_2009}, or tolerability~\citep{blumenthal_electronic_2014} (e.g., not giving excitatory drugs to a patient that has insomnia).  Thus, one might prefer an ML system that agrees with these human-based rules as much as possible.  

Several ML methods have been proposed that combine human expertise in conjunction with training data to perform a prediction task~\citep{mozannar2020consistent, Gennatas4571}.
Some of these methods such as~\citep{madras2018predict} explicitly focus on modeling the \emph{interaction} between an automated ML model and an external decision-maker; the decision-maker determines whether to reject a particular decision made by the model based on the model's confidence and the expertise of the decision-maker. An extension to this procedure in~\citep{mozannar2020consistent} describes when to defer decisions to a downstream decision-maker based solely on samples of the expert's decisions. In contrast to these approaches, we propose a ML system that complements human expertise only when needed, that is, it gives preference to human-based rules as much as possible, subject to explicit  performance constraints in the optimization problem.

In this work, we develop a novel mixture-of-experts (MoE) approach, called \textit{Preferential MoE}, that explicitly incorporates human expertise in learning to provide predictions that align with human-based rules as frequently as possible without losing performance.
The MoE framework allows for an intuitive way to combine ML with clinical expertise.
Importantly, Preferential MoE provides a means of enforcing preference for the human decision rules, as well as an interpretable gating function that allows us to understand when data-driven or clinical expertise should be used. Specifically, we identify when a human decision rule should be followed, and when it makes more sense to provide an alternative data-driven prediction.
Overall, by explicitly incorporating and optimizing for human expertise in our predictions, we obtain models that aligns better with human knowledge, making them easier to inspect, audit and trust.

\section{Related Work}

\paragraph{Human-ML decision making systems.}
There is a long history of approaches to incorporate human expertise in the architecture of ML systems. In particular, \citet{towell_knowledge-based_1994} and \citet{tran_deep_2018} propose methods that map rules to elements of a neural network. \citet{wu_knowledge_2018} incorporates human-based knowledge gates into Recurrent Neural Networks for question answering or text matching. Closer in spirit, \citet{wang_learning_2018} constrained a ML model to be more credible by relying as much as possible on input predictors that are intuitive for human experts. All these approaches include human expertise as input or intermediate features, whereas we assume that the expert information is available in the form of output decision rules, on which we want to rely as much as possible. Recently, \citet{chattha_kinn:_2019} learns a ML system complementary to humans by modeling the residual of humans in the context of timeseries.  Here we focus on classification, and additionally provide an interpretable explanation about when to rely on human-based rules. Finally, \citet{hu_harnessing_2016} proposes a knowledge distillation approach, where human decisions are used as a teacher, and a student network is trained to mimic the human decisions while performing well on test data. Unlike implicitly assuming human expertise as additional ground truth labels (teacher), this work has the capacity of ignoring human rules if those are found unreliable.

\vspace{-0.45cm}
\paragraph{Mixture of Experts.} In the ML community, mixture-of-expert (MoE) models \citep{jacobs1991adaptive,jordan1994hierarchical} are frequently used to leverage different types of expertise in decision-making. The model works by explicitly learning a partition of the input space such that different regions of the domain may be assigned to different specialized sub-models or experts. MoEs have also been applied to several healthcare domains such as HIV \citep{parbhoo2017combining, parbhoo2018improving}. The proposed approach Preferential MoE is different in three regards: first, we explicitly incorporate human knowledge in the form of therapy standards and guidelines for medical decision-making. Second, our framework expresses an explicit preference for a specific expert (human-based), and trains an ML-based expert to complement the primary expert; third, we learn an 
interpretable gating function, which makes the model easy-to-interpret and give us information on when human-based rules should be followed.

\vspace{-0.45cm}
\paragraph{Learning to defer approaches.} \citep{madras2018predict, mozannar2020consistent}  propose MoE classification models to be used as triage tools, where only the most critical decisions are deferred to a medical expert, whilst relying on data-driven approaches the majority of the time. Specifically, these classifiers are trained based solely on the samples of an expert's decisions. Other approaches for integrating human expertise in decision-making such as \citep{raghu2019algorithmic, wilder2020learning} train a standard classifier on the data and subsequently obtain uncertainty estimates based on this classifier and the human expert. The decision is ultimately deferred to the expert with the lowest uncertainty. Unlike triage methods, we view human expertise as \emph{complementary} to data-driven approaches and explicitly leverage these sources of knowledge to inform better predictions. That is, we optimize to rely on human expertise as much as possible, except for those regions for which human-based rules are inadequate. Our training samples consist of generic (potentially partial) rules that have been specified by humans a priori.

\section{Methodology}

\begin{figure}
	\centering
	\includegraphics[width=0.90\linewidth]{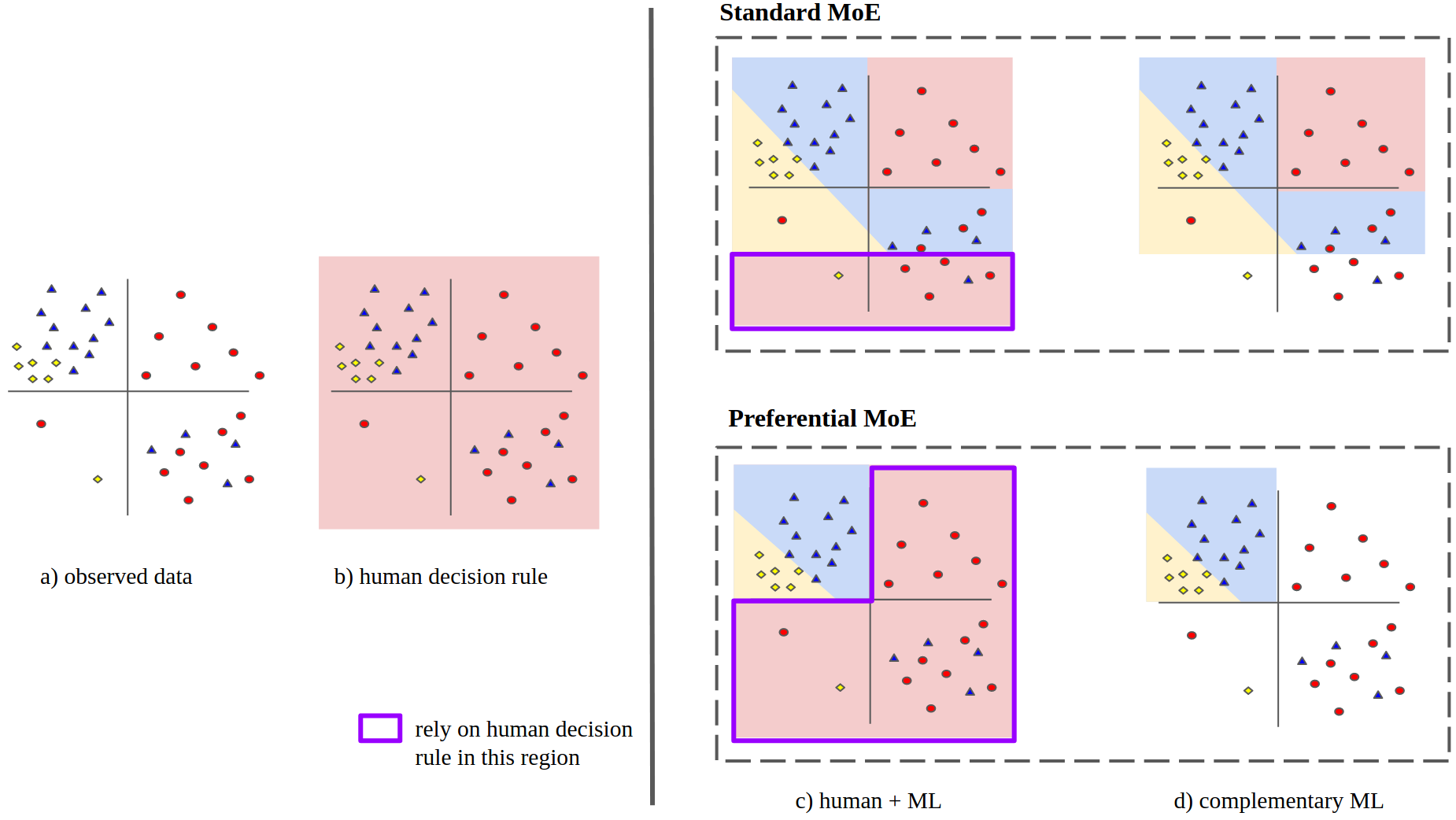}
	\caption{\textbf{Preferential MoE}: a mixture-of-experts (MoE) approach that relies on human decision rules subject to performance guarantees, and only disagrees with humans when a data-driven model can do better. A standard MoE exhibits similar predictive performance, but relies less on human rules. The proposed approach also provides insights on when/why human rules should be followed or not via an interpretable gating function (region highlighted in purple).}
	\label{fig:preferentialmoe}
\end{figure}

In this section, we present Preferential MoE.
The proposed approach fulfills two desiderata.
First, Preferential MoE relies on the human rules as much as possible while preserving predictive performance.
When the human-based rules are damaging w.r.t the prediction task, the proposed approach is able to overrule them (that is, we recover the same solution as the unconstrained standard MoE formulation).
Second, the gating function is interpretable, providing information on when each human guideline is applicable.

An overview of how Preferential MoE operates is illustrated in Figure \ref{fig:preferentialmoe}. Colors in columns b)-d) represent predictive decision boundaries.  In the proposed example, the human-based rule predicts red everywhere in the input space (sketch b). The third column (sketch c) shows the final predictions (colors) for each region of the input space. Each prediction either comes from the human decision rule, or from a data-based ML classifier. We learn a gating function (highlighted in purple) to select which classifier to rely on, as well as a complementary ML classifier to make predictions in regions outside of the purple region. In this diagram, both the standard MoE and the preferential MoE exhibit same predictive performance; however, the preferential MoE relies on humans much more often.

More formally, let $\mathcal{D}=\{(\x_n,y_n)\}^N_{n=1}$ be a dataset of observations where $\x_n \in \mathbb{R}^d$ are the covariates and $y_n \in \{1,...,K\}$ is a categorical outcome for a specific prediction task. Let the \emph{guideline} function $ g : \mathbb{R}^d\rightarrow \{1,0,-1\}$ be an aggregated function encoding all available human decision rules, whose input are the covariates, and the output might be any output category, or a flag ($-1$) indicating that the rule is not applicable (human does not know).
This human guideline function $g$ is fixed a priori by domain knowledge or well-established medical practice. 
In the case of not having access to an explicit human function $g$, but samples of past human decisions instead, we can pre-train a classifier to mimic those human decisions beforehand, and use such classifier as our human-based rules function $g$.

Given dataset $\mathcal{D}$, there might exist several functions that exhibit similar predictive performance, but are qualitatively different.
We want to use expert knowledge (via human-based rules) to guide the optimization such that we are able to find models that have high predictive performance and agree with the human-based rules as much as possible.
In order to accomplish that, we will include both objectives in the proposed optimization.

\paragraph{Modeling.} Our goal is to make predictions that prioritize human-based rules when the data supports (or does not contradict) such knowledge, and learn to defer to another trainable ML expert when the human rules counters empirical evidence.
For that, we propose a new classification model based on a mixture of experts formulation.
Let $ f_\pf : \mathbb{R}^d\rightarrow \Delta^K $ be a trainable ML expert parameterized by $\pf$, where $\Delta^K$ denotes the $(K-1)$-simplex (outcome vectors of $f_\pf$ should sum to one).
Our approach combines the predictions of the ML expert $ f_\pf$ and the human expert $g$ via the gating function $ \rho_\prho : \mathbb{R}^d \rightarrow \{ 0,1 \}$ parametrized by $\prho$; $\rho_\prho$ is another classifier that selects which expert to rely on given the covariates $\x$.
The prediction model of Preferential MoE is formalized as
\begin{equation}\label{eq:ypred-MoE}
\hat{y}_{\theta,w}(\x)=
\begin{cases}
	\:(1-\rho_{w}(\x))f_{\theta}(\x)+\rho_{w}(\x)g(\x) & \quad\text{{if }\;}g(\x)\neq-1\\
	\:f_{\theta}(\x) & \quad\text{{if }\;}g(\x)=-1
\end{cases}
\end{equation}
where $\hat{y}_{\pf,\prho}(\x) \in \Delta^K$, and the likelihood function is given by
\begin{equation}\label{eq:MoE}
	y|\x \sim \Cat\ \Big(\hat{y}_{\theta,w}(\x)\Big).
\end{equation}
Note that the ML expert $ f_\pf $ might make predictions and specialize in input regions where the human rule $ g $ is not applicable or is inaccurate.
In summary, Equation~\eqref{eq:MoE} assumes that every data point $ \x $ can be discriminated by $ f_\pf $ or $ g $, and $ \rho_{\prho} $ makes a deterministic decision on which expert to rely on.
When learning $ \rho_w $, we will prioritize human-based rules $ g$ during inference.
Notice that \eqref{eq:ypred-MoE} produces a non-convex prediction model which may be difficult to optimize.
MoE classification models are notorious for converging to local optima \citep{jordan1995convergence}.

The gating function selects when a decision should rely on human-based rule or a trained expert.
By making $ \rho_w $ an interpretable function, e.g., a linear classifier or a decision tree, the model learns which features are important for human-based decision making, and identifies the regions of other expert classifiers. We note that, even if the gating function $ \rho_w $ is chosen to be interpretable, our approach does not provide theoretical guarantees on identifying all the regions suitable for human decision rules. 
More generally, $ \rho_w $ can also be a non-interpretable function, e.g., a neural network.
In such case, the gating function still identifies regions appropriate for human-based decisions, although it may miss the interpretability of the parameters $ w $.  
Overall, our framework allows model constructions that balance flexibility and interpretability suitable to different applications.

Before continuing, we note that our Preferential MoE differs from other MoE approaches that learn to defer in the following ways.
\citet{madras2018predict} trains multiple experts and learns when to defer to a human, based on accuracy, uncertainty and fairness, via regularization of the loss function.
\citet{mozannar2020consistent} proposes a novel cost sensitive function, studies its theoretical properties, and learns the regions of space where each expert is accurate.
Neither of these approaches prioritize human rules, so it is possible that given an expressive expert, the mixture learns to reject human-based rules and relies solely on the trained expert.
Such deferral may come at the cost of not truly identifying the input regions where human rules are adequate.
Because in many applications identifiability may be critical to understand when humans' decisions are suitable, we propose a new inference procedure that defers primarily to humans over other methods while preserving pre-specified performance guarantees.
Preferential MoE will maximize use of human-based rules with an explicit constraint that the performance cannot be signficantly worse than the standard MoE without preferences.

\vspace{-0.2cm}
\paragraph{Problem formalization.}
Our formulation as an optimization problem needs to reflect the following criteria: (i) we want to minimize the predictive error, (ii) we want to follow human-based rules as frequently as possible without hurting performance.

We optimize for predictive performance by minimizing the cross-entropy $L^{\gamma}_{\theta,w}(\calD)$ with respect to the predictions from Equation~\eqref{eq:ypred-MoE}; this corresponds to a standard maximum log-likelihood estimator for the probabilistic model in Equation~\eqref{eq:MoE} with an additional regularizer.
For example, if the outcome is binary we write
\begin{equation}\label{eq:bce}
	L^{\gamma}_{\theta,w}(\calD) = \sum_{n=1}^N -\left[ y_n\ln(\hat{y}_{\theta,w}(\x_n)) + (1 - y_n) \ln(1 - \hat{y}_{\theta,w}(\x_n))\right] + \gamma ||w||_1,
\end{equation}
where $\gamma \geq 0$ is a regularization weight that controls the trade-off between predictive performance and sparsity of $ w $. 
A sparse $ w $ can help identify important features for the gating function $ \rho_w(x) $.

We bound the cross-entropy loss $L^{\gamma}_{\theta,w}(\calD)$ with a prefixed optimized value for performance guarantees.
Denote $ L^{\gamma}_{\theta^{\ast},w^{\ast}}({\cal D}) $ an attainable loss where $ \theta^{\ast}$ and $w^{\ast} $ are solutions of minimizing $ L^{\gamma}_{\theta,w}(\calD) $ for the stated MoE in Equation~\eqref{eq:MoE}.
Consider a margin $ \varepsilon \geq 0 $ measuring an acceptable performance decrease, and consider the constraint:
\begin{equation}\label{eq:performance}
	L^{\gamma}_{\theta,w}({\cal D}) \leq (1+\varepsilon) L^{\gamma}_{\theta^{\ast},w^{\ast}}({\cal D}).
\end{equation}
Equation \eqref{eq:performance} guarantees that the performance loss will not increase more than specified, and will maintain predictive error results.
We introduce sets $ \Theta \subset \mathbb{R}^q $ and $ W \subset \mathbb{R}^p $ such that $ \theta \in \Theta $ and $ w \in W $.  Variables $ \theta $ and $ w $ do not need to have same dimensions and can be constructed with different model classifiers.

We present next the problem formulation for Preferential MoE:
\begin{equation}\label{eq:game}
\calG:\qquad
\begin{IEEEeqnarraybox}[][c]{c}
\text{(player 1)}\quad
\begin{IEEEeqnarraybox}[][t]{r'l}
\min_{w\in W} & -\smsum_{n=1}^N \ln (\rho_{w}(x)) \\
\text{s.t.} & L^{\gamma}_{\theta,w}({\cal D}) \leq (1+\varepsilon) L^{\gamma}_{\theta^{\ast},w^{\ast}}({\cal D})
\end{IEEEeqnarraybox}
\qquad
\text{(player 2)}\quad
\begin{IEEEeqnarraybox}[][t]{r'l}
\min_{\theta\in \Theta} & L^{\gamma}_{\theta,w}(\calD)
\end{IEEEeqnarraybox}.
\end{IEEEeqnarraybox}
\end{equation}
We refer to \eqref{eq:game} as game $ \calG $.
Using game theory terminology, there are 2 players and each player optimizes their own objective, variables and constraints, while taking into account the other player's decisions.
Notice that $ \calG $ explicitly models our discussed goals: player 1, which is optimizing the gating function, maximizes the number of human-based decisions; player 2, which optimizes the classifier $ f_{\theta} $, minimizes prediction error according to the loss function \eqref{eq:bce}.
Note that the negative logarithm is a monotone transformation that helps obtain a convex objective for player 1.
Player 1 also imposes the performance constraint and limits the classification loss.

$ \calG $ is a particular instance of a generalized Nash equilibrium problem (GNEP) \citep{pang2005quasi}.
Our goal is to minimize both objectives and reach an equilibrium point known as Nash equilibrium, where no player is incentivized to change its decision based on the other player's actions. In general, finding the Nash equilibrium is particularly challenging as a result of the dynamic nature of the feasibility region.

Existence of solutions is guaranteed for $ \calG $ as discussed in \cref{prop:existence} (see \cref{apx:proofs}).
Next, we present two algorithms for solving~\eqref{eq:game}, and discuss their properties and convergence.

\section{Inference Algorithms}
\label{sec:algorithms}

Inference for determining $ \theta $ and $ w $ from $ \calG $ proceeds in two steps:
\begin{enumerate}
	\item \textbf{Unconstrained optimization}: we train a standard MoE model from \cref{eq:MoE} by minimizing the performance loss $L^{\gamma}_{\theta,w}(\calD)$ described in \cref{eq:bce}. This step yields a performance reference  value of $ L^{\gamma}_{\theta^{\ast},w^{\ast}}({\cal D}) $ which we will aim to maintain up to a certain margin $\varepsilon$. We use the optimal parameters $\theta^{\ast}$ and $w^{\ast}$ from the unconstrained problem as warm initialization for the next step.
	\item \textbf{Constrained optimization}: we solve game $ \calG $ initializing from previous solution.
\end{enumerate}
We discuss two algorithms for solving $ \calG $.
The first proposal combines both objectives and uses a log-barrier method to approximate a solution.
The second proposal takes gradient steps that minimize each objectives alternatively and projects to the feasible region.
Both methods have convergence guarantees.

\paragraph{Log-Barrier Method.}

We want to approximate a solution of $ \calG $ by simplifying its formulation.
We move player 1's constraint to the objective using a log-barrier penalty used in interior point methods \citep[Chapter 11]{Boyd2004} and combine both objectives (see \cref{apx:log-barrier}).
These operations transform $ \calG $ into the following unconstrained non-convex optimization problem:
\begin{equation}\label{eq:log-barrier}
	\min_{\theta\in \Theta,w\in W} \quad -t \sum_{n = 1}^N \ln\left(\rho_w(x_n)\right) -\ln\left((1+\varepsilon) L^{\gamma}_{\theta^{\ast},w^{\ast}}({\cal D}) -L^{\gamma}_{\theta,w}({\cal D})\right).
\end{equation}
The first term of equation \eqref{eq:log-barrier} corresponds to player 1's objective, and the second term to the log-barrier function $\widehat{I}(u) = -1/t \ln(-u) $ transforming its constraint, which also aligns with player 2's objective.
Note that a solution for problem \eqref{eq:log-barrier} exists, provided \cref{assm:compactness} in \cref{apx:proofs} holds.

The log-barrier argument is susceptible of becoming negative inside the logarithm and be a source of numerical instability, so care needs to be taken with step sizes and correct initialization (warm-start).
Parameter $ t $ is a hyperparameter that weights the satisfiability of the constraint, and the approximation improves as $ t $ grows.
Note that this approximated form encourages that the difference $L^{\gamma}_{\theta^{\ast},w^{\ast}}({\cal D}) -L^{\gamma}_{\theta,w}({\cal D})$ becomes large, regardless of the constraint already being satisfied.
This has the desirable effect of continuously minimizing $ L^{\gamma}_{\theta,w}({\cal D}) $.
Finally, because of the non-convex nature of the problem, gradient descent methods only guarantee convergence to stationary solutions.

\paragraph{Projected Gradient Method.}
Player 2's decisions affect player 1's constraint, and player 1's affect player's 2 objective in game $ \calG $.
A simple algorithm would be to alternate solving subproblems and repeat until convergence.
Such schemes are only guaranteed to converge under very stringent conditions of monotonicity of the game.
Monotonocity is a desirable property of multivariate mappings, informally stating that a small change in the input guarantees a bounded change in the output, therefore permitting dynamics of control towards stable solutions.
We refer the reader to \citet{scutari2012monotone} for definitions, properties and algorithms for solving monotone games.

We present \cref{alg:game} for solving $ \calG $. 
The algorithm makes a gradient update on each objective, and projects the result onto the feasibility region.
We denote estimates on iteration $ k $ with $ \theta^k $ and $ w^k $.
The feasibility region is denoted with $ K_{\varepsilon}(\theta^{k+1}) $, and is formally introduced in \cref{apx:proofs}. 
The operation $ \Pi_{K_{\varepsilon}(\theta^{k+1})} $ denotes projection of $ w $ onto the set $ K_{\varepsilon}(\theta^{k+1}) $.
The projection operation solves the following optimization problem
\begin{equation}\label{eq:projection}
\begin{IEEEeqnarraybox}[][c]{rCl'l}
	\Pi_{K_{\varepsilon}(\theta^{k+1})} \big( z \big) & = & \argmin_{w\in \big(W \cap K_{\varepsilon}(\theta^{k+1})\big)} & \frac{1}{2} \Vert w - z\Vert ^2,
\end{IEEEeqnarraybox}
\end{equation}
whose solution can be efficiently computed via a bisection search, described in \cref{alg:bisection}.
The optimization inside the while loop in \cref{alg:bisection} can be solved via L-BFGS \citep{liu1989limited}, or other efficient algorithm, since it does not have a known analytical expxression.

We state the convergence of \cref{alg:game} to a Nash Equilibrium in the next theorem.
Proof is provided in \cref{apx:proofs}.

\begin{theorem}[Convergence]\label{th:convergence}
	Assume \cref{assm:compactness,assm:non-empty-set,assm:convexity} are satisfied (see \cref{apx:proofs}), and step-size $ \alpha $ is small enough. Then, $ \calG $ is monotone and \cref{alg:game} converges to a Nash Equilibrium of game $ \calG $.
\end{theorem}

\begin{minipage}{0.48\textwidth}
	\begin{algorithm}[H]
		\SetAlgoLined
		\KwIn{$\calD,\: \varepsilon,\: L^{\gamma}_{\theta^{\ast},w^{\ast}},\:\set{\alpha^k}$}
		\KwOut{$ \theta $ and $ w $.}
		Initialization: $ \theta^{0} \leftarrow \theta^{\ast} $, $ w^0 \leftarrow w^{\ast}$, $ k \leftarrow 0 $ \;
		\While{stopping criteria not satisfied}{
			$ \theta^{k+1} \leftarrow \theta^{k} - \alpha^k \frac{\partial}{\partial \theta} L^{\gamma}_{\theta^k,w^k} $ \;
			$ w^{k+1} \leftarrow  \Pi_{K_{\varepsilon}(\theta^{k+1})} \big( w^{k} - \alpha^k \frac{\partial}{\partial \theta} L^{\gamma}_{\theta^{k+1},w^k} \big) $ \;
			$ k \leftarrow k+1 $
		}
		\caption{Projected Gradient Descent}
		\label{alg:game}
	\end{algorithm}
\end{minipage}
\hfill
\begin{minipage}{0.5\textwidth}
	\begin{algorithm}[H]
		\SetAlgoLined
		\KwIn{$\theta^{k+1},\:\varepsilon\:,z\in\mathbb{R}^p,\: \overline{\lambda} $; Initialization: $ \underline{\lambda}\leftarrow 0 $}
		\KwOut{$ w^{k+1} $.}
		\While{$ (\overline{\lambda} - \underline{\lambda}) \geq \text{tolerance} $}{
			$ \lambda \leftarrow (\overline{\lambda}+ \underline{\lambda})/2$ \;
			$ w \leftarrow \argmin_{w\in W}\quad \frac{1}{2} \Vert w - z\Vert ^2 + \lambda L^{\gamma}_{\theta,w}({\cal D}) $ \;
			\lIf{$ L^{\gamma}_{\theta,w}({\cal D}) - (1+\varepsilon) L^{\gamma}_{\theta^{\ast},w^{\ast}}({\cal D}) > 0  $}
			{ $ \underline{\lambda} \leftarrow \lambda $ }
			\lElse{ $ \overline{\lambda} \leftarrow \lambda $ } 
		}
		\caption{$ \Pi_{K_{\varepsilon}(\theta^{k+1})} $ (bisection search)}
		\label{alg:bisection}
	\end{algorithm}
\end{minipage}

\section{Results}
\label{sec:results}

We compare the performance of Preferential MoE against several baselines for two medical tasks for the treatment of Human Immunodeficiency Virus (HIV), or pharmacological management of Major Depressive Disorder (MDD). Our baselines include using predictions a) based on a human expert alone; b) a logistic regression ML expert alone; c) a standard mixture-of-experts model (standard MoE); d) the learn-to-defer model in \citep{madras2018predict}; and e) a learn-to-defer model from \citep{mozannar2020consistent}.
For the standard MoE and Preferential MoE, we train models either assuming discrete $\rho(x)$ values to begin with, or assuming continuous $\rho(x)$ values and the discretizing at the end, exploring all operating points for the threshold of the gating function. Here we report the latter, which seems to work better in practice.

\paragraph{Hyperparameter selection.} For both prediction tasks, we explore different learning rates for both, the unconstrained and constrained optimization steps, in the range of $\{10^{-4},  10^{-3}, 0.01, 0.1\}$. We also explore a range of regularization parameters $\gamma \in \{0.0, 0.001, 0.01, 0.05, 0.1, 1.0\}$ for the gating function, and select those that maximize predictive performance in a validation set. For the psychiatry dataset, we additionally regularize the ML classifier with an L1 penalty to avoid overfitting due to the high-dimensionality of the input space.
We fix the margin $\varepsilon = 0.1$, and the trade-off parameter $t = 5.0$ for the log-barrier penalty in Equation~\eqref{eq:log-barrier}. Our results were stable to perturbations of these parameters. Intuitively, $t$ can be matched to existing interior-point algorithms and is quite robust with appropriate gradient step sizes. The margin $\epsilon$ affects model´s accuracy, but even if there is no direct mapping from its value to a desired performance level, its impact was similar in the range $\epsilon \in [1e^{-2}, 2e^{-1}]$. Setting $\epsilon$ too small can make the model not move from the initialization point, and its solution stay similar to the standard MoE’s.

\vspace{-0.4cm}
\paragraph{Evaluation metrics.} To evaluate Preferential MoE and other baselines, we measure performance as Area-Under-the-operating-ROC-Curve (AUC), as well as predictive accuracy (percentage of correct predictions) for a fine-grid of threshold values, both for the gating function and final predictions. Note that all thresholds are chosen by cross-validation, we thus guarantee that the right thresholds (w.r.t the most adequate metric for each downstream task) are selected, in a data-driven manner.
We report \textit{coverage} as a measure of how frequently (in percentage) each model relies on the human-based guideline function $g$. More specifically, we define \textit{soft-coverage} and \textit{hard-coverage}$(t)$ for a given gating function threshold $t$ as follows:
\begin{equation}
\text{soft-coverage} = 100.0 \times \mathbb{E}[\rho(x) ] \quad \quad \text{hard-coverage}(t) = 100.0 \times \mathbb{E}\big[\mathbbm{1}[\rho(x)\geq t] \big].\label{eq:coverage}
\end{equation}

\subsection{Human Immunodeficiency Virus (HIV) Therapy Outcome Prediction}
HIV currently affects more than 36 million people worldwide. The life-long use of combinations of antiretrovirals has largely helped combat the virus in most parts of the world and has transformed the virus from a life-threatening condition to a chronic illness. However, administering therapies is tricky as patients frequently suffer from drug resistance, viral relapses or spikes, as well as adherence issues and several other side-effects from use of antiretrovirals.

We identified individuals between 18-72 years of age from the EuResist database comprising of genotype, phenotype and clinical information of over 65\,000 individuals in response to antiretroviral therapy administered between the years 1983 and 2018. We focus on a subset of 36\,780 of these patients who received at least 3 prior treatments and base our predictions on the genotype, phenotype, clinical and demographic information of these individuals. The curated dataset contains a total of 384 such features. Our goal is to predict short-term therapy success where viral suppression is maintained for at least 40 days after a therapy is administered. 

\begin{table}[b]
	\centering
	\vspace{-0.3cm}
	\begin{tabular}{c|cccc}\toprule
		\midrule
		& \multicolumn{2}{c}{AUC} & \multicolumn{2}{c}{soft coverage (\%)} \\
		\multicolumn{1}{c|}{Baselines} & mean & CI & mean & CI \\
		\midrule
		ML only & 0.64 & [0.63-0.65] & 0.00 & [0.00-0.00] \\
		Learn-to-defer\citep{madras2018predict} & 0.71 & [0.68-0.72]& 54.07 & [48.18 - 55.63]\\
		Consistent Learn-to-defer \citep{mozannar2020consistent} &0.66 &[0.62-0.69] & 56.81 &[50.02 - 57.62]\\
		Standard MoE (unconstrained) & 0.69 & [0.69-0.70] & 52.87 & [51.19-54.55] \\
		\midrule
		Preferential MoE (log barrier) & \textbf{0.74} & [0.72-0.76] & 62.06 & [60.8-63.32] \\
		Preferential MoE (projected gradient) & \textbf{0.74} & [0.73-0.75] & \textbf{63.18} & [61.7-64.66] \\
		\bottomrule
	\end{tabular}
	\vspace{-0.1cm}
	\caption{\label{tab:table1hiv} \textbf{Performance vs Coverage (HIV)}:  Preferential MoE relies much more often on human expertise while preserving predictive performance. Predictive performance measured by Area-Under-the-operating-ROC-Curve (AUC); Reliance on human decision rules based on soft coverage, as defined in Equation~\eqref{eq:coverage}.}
\end{table}

\begin{figure}
	\centering
	\begin{minipage}{0.50\textwidth}
		\centering
	\vspace{0.8cm}
	\includegraphics[width=1.05\linewidth]{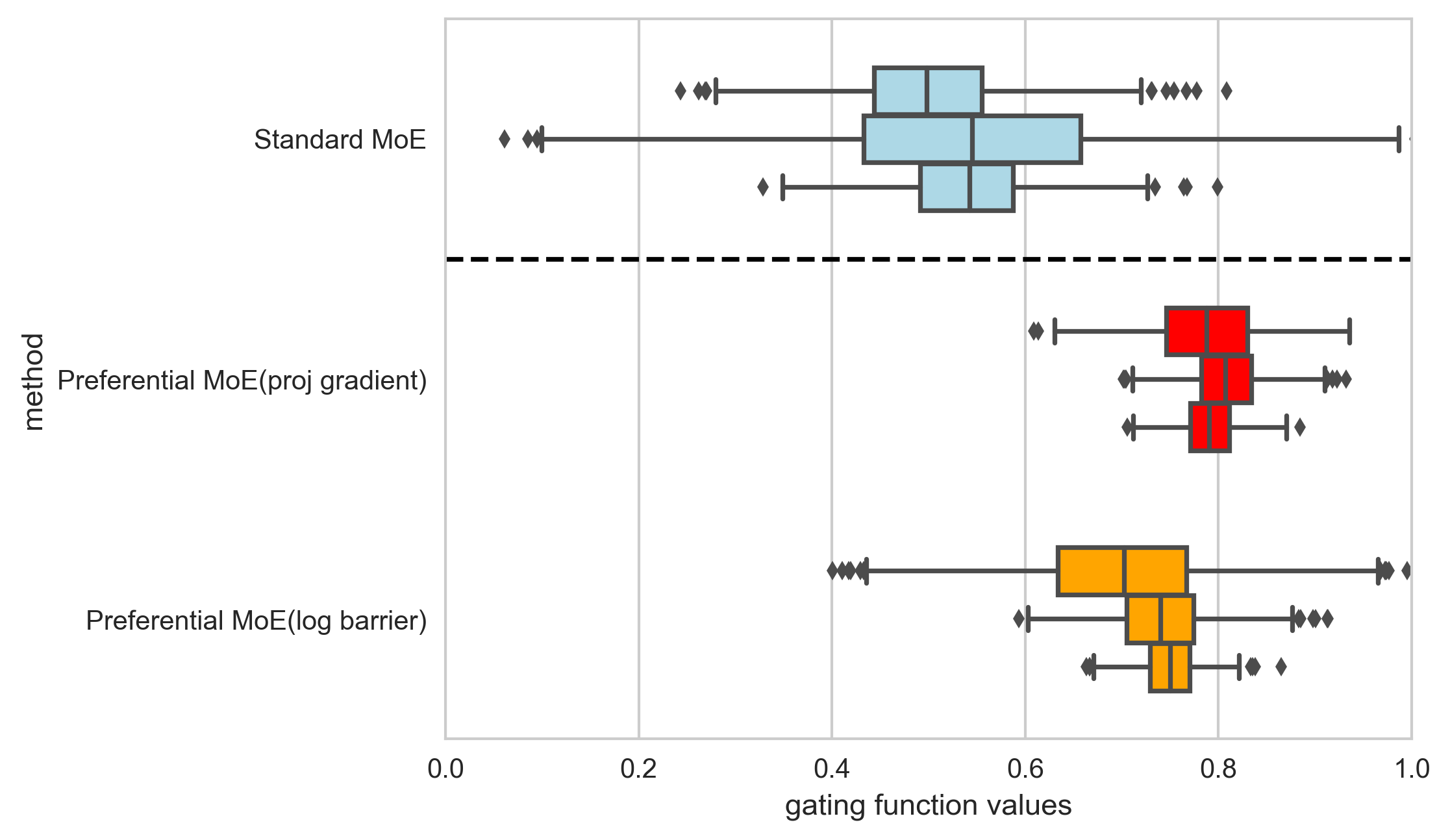}
	\label{fig:histogramhiv}
	\vspace{-0.2cm}
	\caption{\textbf{$\boldsymbol{\rho(x)}$ values in the test set for HIV.} Preferential MoE pushes up the values for the gating functions, favoring human decision rules more frequently in the input space. Each box plot corresponds to a different random seed (we report 3 different initializations per method).}
	\label{fig:hivrho}
	\end{minipage}\hfill
	\begin{minipage}{0.46\textwidth}
		\centering
	\includegraphics[width=0.78\linewidth]{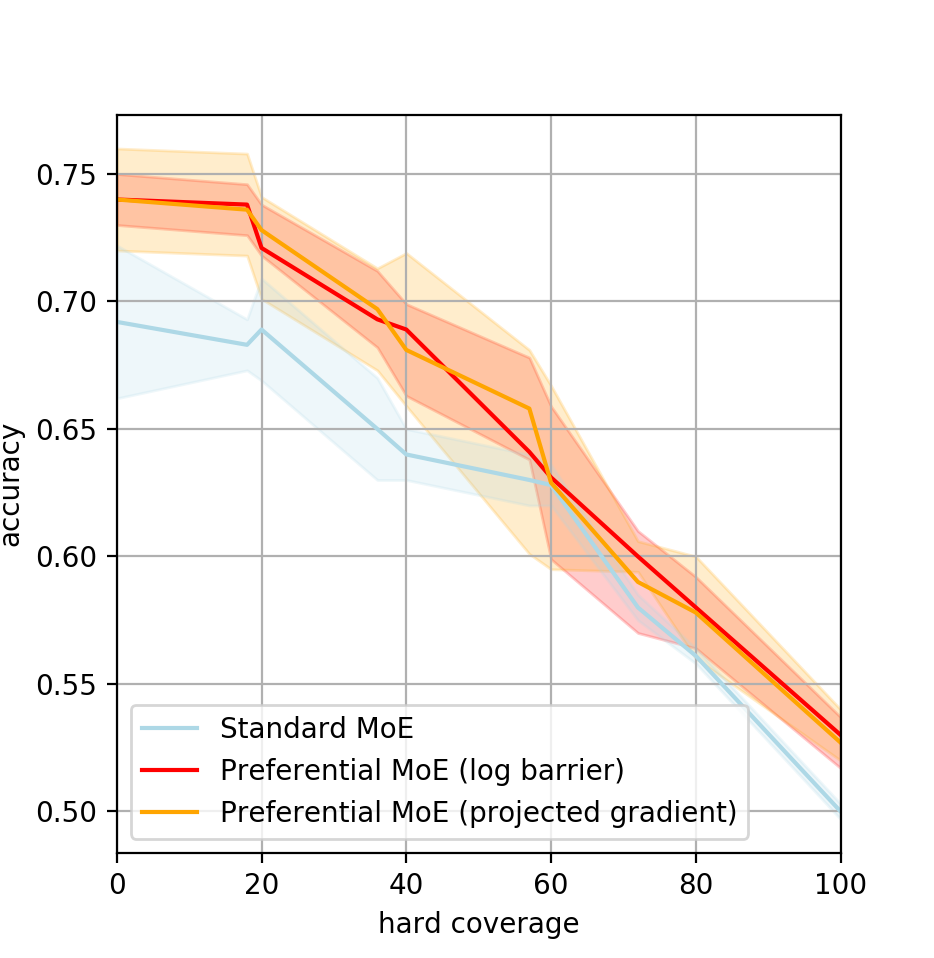}
	\vspace{-0.2cm}
	\caption{\textbf{Accuracy-coverage trade-off.} Preferential MoE (trained by the log barrier or projected gradient method) for HIV either relies more on human rules for the same predictive accuracy, or gets higher accuracy for the same coverage with human rules.}
\label{fig:curvehiv}
	\end{minipage}
\end{figure}

Table~\ref{tab:table1hiv} shows predictive performance and coverage results for the proposed approach and competing baselines. Compared to other approaches, Preferential MoE exhibits highest soft coverage while either retaining or improving predictive performance.
Figure~\ref{fig:curvehiv} compares the accuracy relative to hard thresholding of the coverage for each of the MoE models. In the HIV setting, both variants of the Preferential MoE outperform the standard MoE approach at various coverage values. At 60\% coverage, the methods all seem to perform relatively similarly in terms of accuracy.

Importantly, Preferential MoE allows us to incorporate human expertise into the prediction task and provides us with insights of when it makes sense to follow the rules based on the gating function. Table ~\ref{tab:predictorshiv} provides a sparse list of predictors and corresponding weights averaged over 10 random seeds for the gating function. These predictors are associated with regions where it makes sense to follow human intuition. While Standard MoE identifies blood count data, certain mutations and a patient's risk group as meaningful factors, Preferential MoE identifies a significantly different set of predictors. Notably, many of the predictors identified in the latter correspond to cases where patients have additional conditions such as lipodystrophy or side effects to medication where it is preferable to rely on human judgement to determine how to treat these individuals. Figure~\ref{fig:hivrho} compares the gating function values $\rho(x)$ in the test set for HIV. Unsurprisingly, Preferential MoE shows a higher preference for relying on human rules.


\definecolor{Gray}{gray}{0.93}
\definecolor{ao}{rgb}{0.0, 0.5, 0.0}
\definecolor{applegreen}{rgb}{0.55, 0.71, 0.0}

\begin{table}[h!]
	\small
	\centering
	\begin{minipage}[t]{0.45\linewidth}
		\begin{tabular}{cl}
			\toprule
			\toprule
			\textbf{Weight} $w$ & \textbf{Covariate Description} \\
			\midrule
			+0.1612 $\pm$ 0.014 &  CD8${+}$ cell count (cells/ml) \\
			\rowcolor{Gray}
			\color{red} -0.1161 $\pm$ 0.002 & \color{red} Reverse Transcriptase Mutation 67N  \\
			-0.0310 $\pm$ 0.025 & Protease Mutation 20M \\
			\rowcolor{Gray}
			\color{red} 0.0280 $\pm$ 0.001 & \color{red} Blood count; complete (CBC)  \\
			-0.0195 $\pm$ 0.005 & Co-infection of Hepatitis C\\
			-0.0156 $\pm$ 0.001 & Stavudine \\
			-0.0124 $\pm$ 0.011 & Reverse Transcriptase Mutation 215YF \\
			\rowcolor{Gray}
			\color{red} +0.0121 $\pm$ 0.020 & \color{red} Nevirapine \\
			\rowcolor{Gray}
			\color{red}-0.0068 $\pm$ 0.031 & \color{red} Risk group MSM \\
			\rowcolor{Gray}
			\color{red} -0.0055 $\pm$ 0.005 & \color{red} Age  \\
			\bottomrule
		\end{tabular}
	\end{minipage}
	\qquad \quad
	\begin{minipage}[t]{0.45\linewidth}
		\begin{tabular}{cl}
			\toprule
			\toprule
			\textbf{Weight} $w$ & \textbf{Covariate Description} \\
			\midrule
			\rowcolor{Gray}
			\color{ao} +0.0359 $\pm$ 0.022 & \color{ao} CD4 + cell count (cells/ml) \\
			-0.0236  $\pm$ 0.027 & Baseline Viral Load \\
			\rowcolor{Gray}
			\color{ao} +0.0151 $\pm$ 0.030 & \color{ao} High Adherence \color{ao}  \\
			+0.0150 $\pm$ 0.001 & Number of Prior Treatment Lines \\
			\rowcolor{Gray}
			\color{ao} +0.076 $\pm$ 0.007 & \color{ao} Pregnancy \\
			-0.0055 $\pm 0.016$ & Reverse Transcriptase Mutation 184V\\
			-0.0035 $\pm 0.002$ & Race black \\
			-0.0026 $\pm 0.001$ & Lamivudine  \\
			\rowcolor{Gray}
			\color{ao} +0.0025 $\pm$ 0.003 & \color{ao} Anaemia \\
			\rowcolor{Gray}
			\color{ao} +0.0012 $\pm$ 0.007 & \color{ao} Lipodystrophy \\
			\bottomrule
		\end{tabular}
	\end{minipage}
	\caption{\label{tab:predictorshiv}\textbf{Interpretation of gating function (HIV)}. Sparse list of predictors describing the regions where human decision rules are followed. We report weight parameters averaged across 10 different random seeds, and for regularization $\gamma$=0.1). (left) Standard MoE (predictors after step 1 in training); (right) preferential MoE (predictors after step 2 in training). Highlighted in red/green are those predictors that disappear/pop-up after step 2 in training.}
\end{table}

\subsection{Prediction of Antipsychotic for Major Depressive Disorder (MDD)} 
Antidepressant prescription for MDD often involves trial
and error. Roughly 2/3 of individuals diagnosed with MDD do not yield remission with their initial treatment, and 1/4 of patients is expected to dropout against clinical advice before finishing their treatment~\citep{hughes_assessment_2020, pradier_predicting_2020}. The list of potential side-effects translates in tolerability and safety concerns that need to be taken into account while prescribing antidepressants. Here we focus on predicting prescription of antipsychotics, which is a class of medication primarily used to manage psychosis, but often used as an adjunctive treatment in the pharmacological management of MDD. The guideline function $g$ for this prediction task is as follows: if the patient has anxiety or insomnia, promote antipsychotic (predict positive label), if the patient has overweight, avoid antipsychotic (predict negative label).

We identified individuals age 18-80 years drawn from the outpatient clinical networks of two academic medical centers in New England, Massachusetts General Hospital and Brigham and Women’s Hospital. These patients had received at least one electronically-prescribed antidepressant between March 2008 and December 2017 with a diagnosis of MDD or depressive disorder at the nearest visit to that prescription.
The goal is to predict prescription of antipsychotic based on demographic information (gender, race) as well as diagnostic and procedure codes.
Race and gender were self-identified features and were included as a proxy for socio-economic variables.
The curated dataset consists of 3,865 individuals and 1,680 features.

Table~\ref{tab:table1} shows predictive performance and soft coverage results for the proposed approach and competing baselines, averaged across 5 random initializations. We encountered issues training the Learn-to-defer approaches to this data (probably due to its high-dimensionality), so we only include the other baselines. Preferential MoE exhibits highest soft coverage (reliance on human rules) while maintaining (or even slightly improving) predictive performance. 

\begin{table}[h!]
	\centering
	\begin{tabular}{c|cccc}\toprule
		\midrule
		& \multicolumn{2}{c}{AUC} & \multicolumn{2}{c}{soft coverage (\%)} \\
		\multicolumn{1}{c|}{Baselines} & mean & CI & mean & CI \\
		\midrule
		ML only & 0.70 & [0.69-0.71] & 0.00 & [0.00-0.00] \\
		Standard MoE (unconstrained) & 0.71 & [0.70-0.71] & 31.41 & [28.48-34.56] \\
		\midrule
		Preferential MoE (log barrier) & \textbf{0.72} & [0.71-0.73] & \textbf{48.34} & [46.24-51.74] \\
		Preferential MoE (projected gradient) & 0.72 & [0.71-0.72] & 45.06 & [42.85-46.70] \\
		\bottomrule
	\end{tabular}
	\caption{\label{tab:table1} \textbf{Performance vs Coverage (Psychiatry)}:  Preferential MoE relies much more often on human expertise while preserving predictive performance. Predictive performance measured by Area-Under-the-operating-ROC-Curve (AUC); Reliance on human decision rules based on soft coverage, as defined in Equation~\eqref{eq:coverage}.}
\end{table}

Preferential MoE gives us additional information on when to follow such human rules by inspecting the gating function. Table~\ref{tab:predictors} presents the sparse list of predictors for the gating function, associated to regions where human decision rules are followed. By regularizing the gating function classifier with an L1-penalty, we get concise list of predictors to describe those regions. The list on the left correspond to Standard MoE (unconstrained optimization), and the list on the right correspond to Preferential MoE (constrained optimization maximizing reliance on humans).
In both lists, most predictors corresponding to general patient care (examination, hospital care, etc) are negatively-correlated: this can be interpreted as higher reliance on humans in the absence of patient care related codes. 
In the case of Preferential MoE, additional covariates coding for cardiovascular risk factors (highlighted in green) are positively-correlated with reliance on human rules. Such information can be used to explore refinements of the human-based rules.

\begin{figure}
	\centering
	\begin{minipage}{0.54\textwidth}
		\centering
		\includegraphics[width=0.89\linewidth]{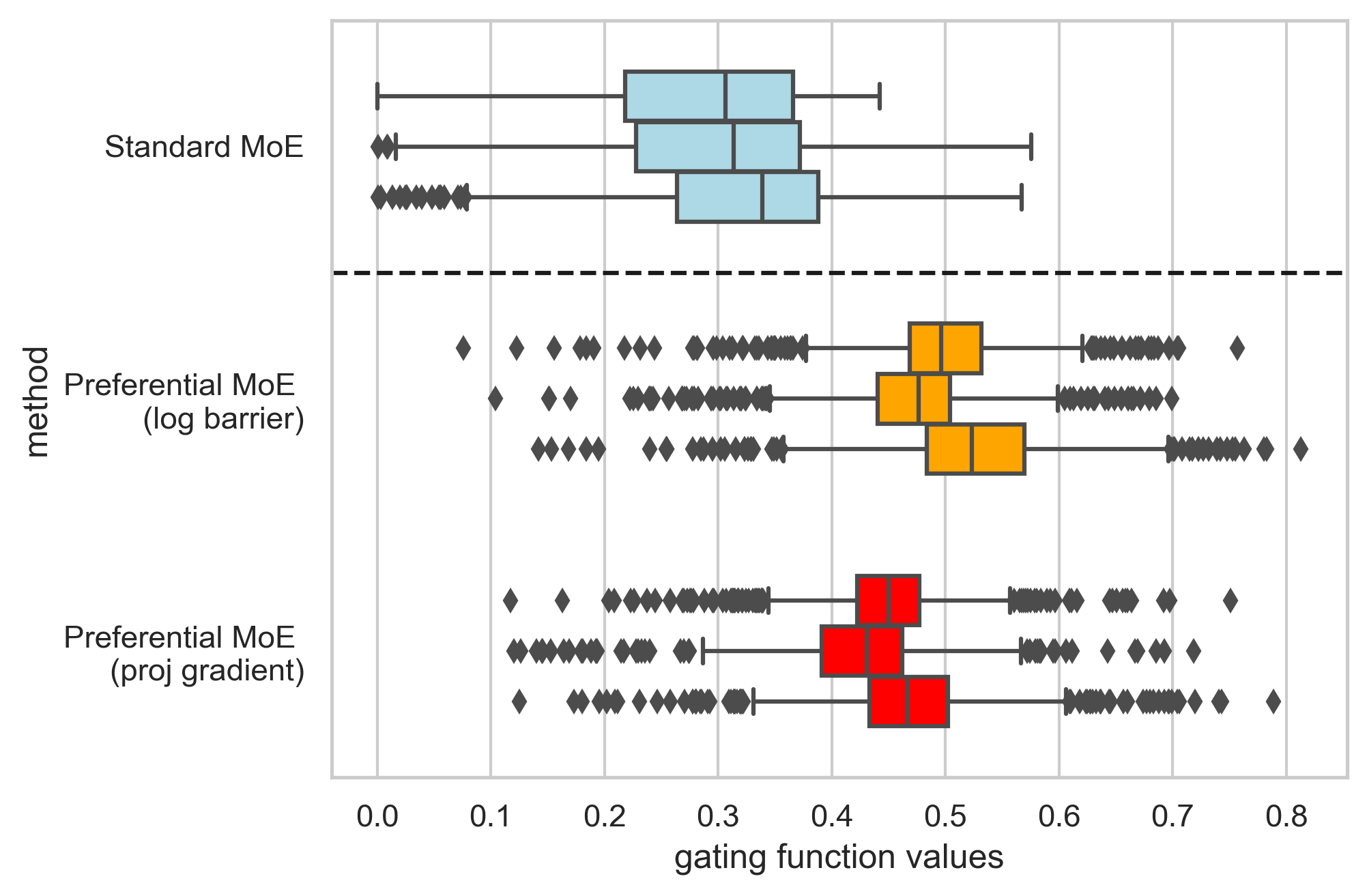}
		\vspace{-0.1cm}
		\caption{\label{fig:hist}\textbf{Histograms for $\boldsymbol{\rho(x)}$ in the test set.} Preferential MoE pushes up the values for the gating function, favoring relying on human decision rules more frequently. Each box plot corresponds to a different random seed (3 per method).}
	\end{minipage}\hfill
	\begin{minipage}{0.43\textwidth}
		\centering
		\includegraphics[width=0.77\linewidth]{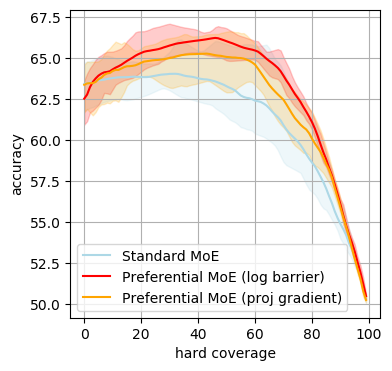}
		\vspace{-0.1cm}
		\caption{\textbf{Accuracy-coverage trade-off.} Preferential MoE either relies more on human rules for the same predictive accuracy, or gets higher accuracy for the same coverage with human rules.}
		\label{fig:curve}
	\end{minipage}
\vspace{0.2cm}
\end{figure}


\definecolor{Gray}{gray}{0.93}
\definecolor{ao}{rgb}{0.0, 0.5, 0.0}
\definecolor{applegreen}{rgb}{0.55, 0.71, 0.0}

\begin{table}[h!]
	\small
	\centering
	\begin{minipage}[t]{0.45\linewidth}
		\begin{tabular}{cl}
			\toprule
			\toprule
			\textbf{Weight} $w$ & \textbf{Covariate Description} \\
			\midrule
-0.0303 $\pm$ 0.0143 & Subsequent hospital care \\
-0.0242 $\pm$ 0.0152 & MDD, recurrent episode \\
-0.0235 $\pm$ 0.0126 & Psychiatric examination \\
-0.0208 $\pm$ 0.0071 & Depressive disorder \\
-0.0153 $\pm$ 0.0106 & Anxiety state \\
\rowcolor{Gray}
\color{red} -0.0117 $\pm$ 0.0089 & \color{red} Office or outpatient visit \\
\rowcolor{Gray}
\color{red} -0.0083 $\pm$ \color{red} 0.0033 & \color{red} Radiologic examination  \\
-0.0073 $\pm$ 0.0062 & Trazodone \\
\rowcolor{Gray}
\color{red} -0.0068 $\pm$ 0.0049 & \color{red} Emergency department visit  \\
\rowcolor{Gray}
\color{red} -0.0031 $\pm$ 0.0062 & \color{red} race white \\
			\bottomrule
		\end{tabular}
	\end{minipage}
	\qquad \quad
	\begin{minipage}[t]{0.45\linewidth}
		\begin{tabular}{cl}
			\toprule
			\toprule
			\textbf{Weight} $w$ & \textbf{Covariate Description} \\
			\midrule
-0.0202 $\pm$ 0.0043 & Subsequent hospital care \\
-0.0115 $\pm$ 0.0075 & MDD, recurrent episode \\
-0.0110 $\pm$ 0.0107 & Psychiatric examination \\
\rowcolor{Gray}
\color{ao} 0.0103 $\pm$ 0.0041 & \color{ao} Office or outpatient visit \\
-0.0070 $\pm$ 0.0112 & Depressive disorder \\
\rowcolor{Gray}
\color{ao} 0.0068 $\pm$ 0.0022 & \color{ao} General medical examination \\
\rowcolor{Gray}
\color{ao} 0.0061 $\pm$ 0.0022 & \color{ao} Type II diabetes \\
\rowcolor{Gray}
\color{ao} 0.0037 $\pm$ 0.0016 & \color{ao} Hypertension \\
-0.0035 $\pm$ 0.0046 & Anxiety state \\
-0.0034 $\pm$ 0.0083 & Trazodone \\
			\bottomrule
		\end{tabular}
	\end{minipage}
	\caption{\label{tab:predictors}\textbf{Interpretation of gating function}. Sparse list of predictors describing the regions where human decision rules are followed. We report weights averaged across 10 different random seeds, and for a regularization parameter $\gamma$=0.1). (left) Standard MoE (predictors after unconstrained step 1 in training); (right) Preferential MoE (predictors after step 2 in training). Highlighted in red/green are those predictors that disappear/pop-up after step 2 in training.}
\end{table}

Figure~\ref{fig:hist} compares the histogram of the gating function values $\rho(x)$ in the test set. As expected, Preferential MoE pushes those values up, reflecting a preference for relying on human rules when possible. Although these values are continuous, we can discretize them using a specific threshold $v$ calibrated on the validation set. Each threshold $v$ yields a different trade-off between accuracy and coverage. Figure~\ref{fig:curve} shows the trade-off between accuracy and hard coverage reachable by these models. As a reference point, the human decision rules have an accuracy of 49.87\% for this prediction task. The curves are averaged over 10 different random seeds, each curve is obtained by changing the thresholds for the gating function and final decision. Overall, Preferential MoE is able to reach better trade-offs, either better accuracy for a given fixed hard coverage, or more hard coverage for a given accuracy level. 


\vspace{-0.2cm}
\section{Limitations}
\vspace{-0.1cm}

First, human guidelines may not be available for specific applications directly.
In such cases, an auxiliary classifier can be trained to predict human labels, as done in~\cite{ madras2018predict,mozannar2020consistent}.
Such a classifier can then be used as a proxy for a human expert in the MoE.
Second, MoEs suffer from local optima, and our methods may converge to different sets of experts and gating functions, depending on initialization.
We advice running multiple initializations and average results (as reported in Section~\ref{sec:results}) or select the best runs based on performance.
Third, our method relies on data that might reflect societal biases, and it may thus suffer from these undesired effects. This is not a limitation specific to our approach; further analysis and exploration is actively pursuit by the community. Finally, we empirically showed that human rules were most useful in regions where data was less prevalent and predictions were less accurate, but our approach does not provide theoretical guarantees in that regard; we let this as future work.

\vspace{-0.2cm}
\section{Conclusion}
\vspace{-0.1cm}

We presented \emph{Preferential MoE}, a mixture of experts that learns and combines a ML general classifier with a human expert, prioritizing the human-based rules.
We presented a game formulation of two objectives, which we solve by a log-barrier method or alternating projected gradient descent.
We evaluate both approaches in the prediction of HIV therapy success, and prescription of antipsychotic for MDD.
Both algorithms preserve performance and maximize coverage of human-based decisions compared to other baselines,  assuming soft and hard decision assignments of the gating function.
Future work will further explore other MoE formulations balancing performance and global optimality of the MoE formulation.

\makeatletter
\renewcommand{\@biblabel}[1]{\hfill #1.}
\makeatother

\bibliographystyle{plainnat}
\bibliography{expert-augmented-machine-learning}

\newpage
\appendix

\section{Derivation of Log-Barrier Method:}
\label{apx:log-barrier}

We derive the log-barrier algorithm presented in \cref{sec:algorithms} first by approximating player 1's subproblem into an unconstrained problem using interior point methods \citep[Chapter 11]{Boyd2004}.
\begin{equation}\label{eq:player1-transform}
	\text{(player 1):}\quad\min_{w\in W} \quad -t \sum_{n = 1}^N \rho_w(x_n) -\ln\left((1+\varepsilon) L^{\gamma}_{\theta^{\ast},w^{\ast}}({\cal D}) -L^{\gamma}_{\theta,w}({\cal D})\right).
\end{equation}

We transform player 2's objective by taking negative logarithm:
\begin{equation}\label{eq:player2-transform}
	\text{(player 2)}\quad \min_{\theta \in \Theta} \quad -\ln\left((1+\varepsilon) L^{\gamma}_{\theta^{\ast},w^{\ast}}({\cal D}) - L^{\gamma}_{\theta,w}({\cal D})   \right)
\end{equation}
Both player 2's objective in \eqref{eq:game} and \eqref{eq:player2-transform} are equivalent because the operation performed is monotone, and \eqref{eq:player2-transform} remains convex in $ \theta $ \citep[Equation 3.10]{Boyd2004}.
Even though we enforced that $L^{\gamma}_{\theta,w}({\cal D}) \leq (1+\varepsilon) L^{\gamma}_{\theta^{\ast},w^{\ast}}({\cal D})$ in the objective, we know that such solution is non-empty (see \cref{assm:non-empty-set}).

After transforming both player's objective from \eqref{eq:game}, we can optimize a Pareto solution that combines both objectives and minimizes all variables.
Since both subproblems have common terms, they simplify and yield a similar formulation as the one presented in \cref{eq:log-barrier}.
Therefore, the log-barrier method searches for a Pareto solution of game $ \calG $ approximately, using interior-point methods to enforce player 1's constraint.

\section{Assumptions and Proofs}
\label{apx:proofs}

\paragraph{Assumptions and existence of solutions.}

Game $ \calG $ models complex relations between variables in both objectives and constraints. We make some simplifying assumptions and provide sufficient guarantees to establish existence of solutions.
These assumptions will also permit to develop algorithms that attain such solutions.

\begin{assumption}\label{assm:compactness}
	Sets $ \Theta $ and $ W $ are compact, convex and non-empty.
\end{assumption}

\Cref{assm:compactness} is mild and practical, constraining that $ \theta $ and $ w $ will take finite values, and that the constraint sets of the optimization procedure will be convex.
Compactness is also a necessary requirement.
Consider for example a binary classification problem where all points are from a single class type, and all points are located in a bounded region of space. 
A logistic regression classifier would allocate all points on one side of the hyperplane, and then maximize its distance from the points without bound and goes to infinity and a solution would not exist. Compactness prevents this undesirable behavior.

Define $  K_{\varepsilon}(\theta) = \set{ w\in W \:|\: L^{\gamma}_{\theta,w}(\calD) \leq (1+\varepsilon) L^{\gamma}_{\theta^{\ast},w^{\ast}}(\calD)} $.
$ K_{\varepsilon}(\theta) $ is a set over $ w $ given some $ \theta\in\Theta $.

\begin{assumption}\label{assm:non-empty-set}
	$ K_{\varepsilon}(\theta) $ is non-empty for every $ \theta \in \Theta $.
\end{assumption}
\cref{assm:non-empty-set} is technical, whose purpose is to guarantee that a solution exists (see \cref{prop:existence} below).
This assumption can always be fulfilled, since $ \Theta = \set{\theta^{\ast}} $ and $ W=\set{w^{\ast}} $ satisfy it by construction, although more interesting problems arise when solving over larger sets.
In practice, the iterates obtained from an alternating optimization process over the player's subproblems in $ \calG $ will satisfy \cref{assm:non-empty-set} at every iteration, and no subproblem will be empty.

\begin{assumption}\label{assm:convexity}
	Functions $ f_{\theta}(x) $ and $ \rho_w(x) $ are log-concave.
\end{assumption}
\Cref{assm:convexity} guarantees that both subproblems of $ \calG $ are convex and can be solved optimally on each variable.
Logistic regression classifiers satisfy the assumption, i.e., $ f_{\theta}(x) = \sigma\big( \theta^T x \big) $ and $ \rho_{w}(x)=\sigma\big( w^T x \big) $, where $ \sigma(z)=1/(1+\exp(-z)) $ is the sigmoid function.
We used these functions to illustrate our results in \cref{sec:results}.

\begin{proposition}\label{prop:existence}
	Given \cref{assm:compactness,assm:non-empty-set,assm:convexity}, a Nash Equilibrium of game $ \calG $ exists.
\end{proposition}
\Cref{prop:existence} guarantees that $ \calG $ is well posed and have a solution.
The proof is given below.

\paragraph{Proof of \cref{prop:existence}.} 

Existence of solution for game $ \calG $ can be established by \citep[Theorem 2]{pang2005quasi}.
Convexity of the set $ K_{\varepsilon}(\theta) $ is guaranteed by construction (the cross-entropy loss is a convex function of $ w $ for fixed $ \theta $).
By \cref{assm:non-empty-set} the set is feasible and non-empty. By \cref{assm:compactness} on sets $ W $ and $ \Theta $ the sets are compact and non-empty.
\cref{assm:convexity} guarantees each subproblem is convex.
Finally, the constraint satisfies the necessary constraint qualifications for dual variables to exist, and a Nash Equilibrium solution must exist.

We state the following result regarding the monotonicity of game $ \calG $:

\begin{proposition}[Monotonicity]\label{prop:monotonicity}
	Given \cref{assm:compactness,assm:non-empty-set,assm:convexity}.
	Assume that $ f_{\theta}(x) $ and $ \rho_w(x) $ are strongly log-concave functions.
	Then, $ \calG $ is monotone.
\end{proposition}

\Cref{prop:monotonicity} is an important intermediate result that allows us to develop \cref{alg:game} and prove its convergence.
The strong log-concavity requirement may be relaxed, see the proof below.
	
\paragraph{Proof of \cref{prop:monotonicity}}

The monotonicty of the game can studied via an equivalent variational inequality (VI) derived from the game subproblems \citep[Proposition 4.1]{scutari2012monotone}.
First, notice that game \eqref{eq:game} can be transformed via Lagrangian relaxation of the first player's subproblem into the following three player game:
\begin{equation}\label{eq:3-player-game}
\begin{IEEEeqnarraybox}[][c]{l'l'l}
	(player 1) & \min_{w\in W} & -\smsum_{n=1}^N -c \ln(\rho_w(x)) + \lambda L^{\gamma}_{\theta,w}({\cal D}) \\
	(player 2) & \min_{\theta \in \Theta} & (1+\lambda) L^{\gamma}_{\theta,w}({\cal D}) \\
	(player 3) & \min_{\lambda \geq 0} & -\lambda \big( L^{\gamma}_{\theta,w}({\cal D}) - (1+\varepsilon) L^{\gamma}_{\theta^{\ast},w^{\ast}}({\cal D})  \big).
\end{IEEEeqnarraybox}
\end{equation}

The third suproblem of \eqref{eq:3-player-game} is derived from the complementary slackness condition of player 1, which has added the constraint multiplied by the dual variable $ \lambda $.
The second player multiplied its objective by positive constant $ (1+\lambda) $, where $ \lambda\geq 0 $, that does not modify the problem.
Neither does $ c $.
Clearly, game \eqref{eq:3-player-game} is equivalent to $ \eqref{eq:game} $.

Game \eqref{eq:3-player-game} has three unconstrained subproblems, so we can now construct an equivalent VI.
Following \citep[Proposition 4.1]{scutari2012monotone}, we obtain $ \VI(\bF,\calQ) $ with $ \bF : \mathbb{R}^{p+ q+1 } \rightarrow \mathbb{R}^{p+ q+1 } $, $ \bF = ((\nabla_w f_1)^T, (\nabla_{\theta} f_2)^T, (\nabla_{\lambda} f_3)^T )^T $, $ f_i $ correspond to the objective function of players $ i \in\set{1,2,3}$ in \cref{eq:3-player-game}; and $ \calQ = W\times \Theta\times\mathbb{R}_{+} $.

Monotonicity can be ascertained by establishing that the symmetric part of the Jacobian of $ \bF $ is positive semidefinite \citep[Section 4.2.3]{scutari2012monotone}.
Indeed,
\begin{equation}\label{eq:jacobian}
	J \bF = 
	\begin{pmatrix}
		c r_{ww} & \lambda \frac{\partial^2}{\partial w \partial\theta} L^{\gamma}_{\theta,w}({\cal D}) & \frac{\partial}{\partial w} L^{\gamma}_{\theta,w}({\cal D}) \\
		(1+\lambda) \frac{\partial^2}{\partial\theta \partial w } L^{\gamma}_{\theta,w}({\cal D}) & (1+\lambda) r_{\theta\theta} & \frac{\partial}{\partial \theta} L^{\gamma}_{\theta,w}({\cal D}) \\
		-\frac{\partial}{\partial w} L^{\gamma}_{\theta,w}({\cal D}) & -\frac{\partial}{\partial \theta} L^{\gamma}_{\theta,w}({\cal D}) & 0
	\end{pmatrix},
\end{equation}
where $ r_{ww} \succ 0 $ and $ r_{ww} \succ 0 $ because of the strong convexity assumption on the objectives.
Notice that the third row and column are antisymmetric, and those terms cancel when studying positive semidefiniteness of $ J\bF $.
Therefore, we only need to show that the upper left corner is positive semi-definite.

We compute the Schur complement of (symmetric) $ J\bF $ \citep[Section A.5.5]{Boyd2004}, i.e.,
\begin{equation}\label{eq:schur}
	S = r_{\theta\theta}-\frac{1+2\lambda}{2c} \Big(\frac{\partial^2}{\partial w \partial\theta} L^{\gamma}_{\theta,w}({\cal D})\Big)^T r_{ww}^{-1} \Big(\frac{\partial^2}{\partial w \partial\theta} L^{\gamma}_{\theta,w}({\cal D}) \Big)
\end{equation}
$ J\bF $ is positive semi-definite if $ S $ is positive definite.
The maximum value of $ \lambda $ is bounded, because $ W $ and $ \Theta $ are compact.
Therefore, there exists $ c>0 $ that necessarily makes $ S \succeq 0 $, proving that $ J\bF $ is positive semi-definite and that game \eqref{eq:3-player-game} is monotone.
Because \eqref{eq:3-player-game} and \eqref{eq:game} are equivalent, both games are monotone.

The strong convexity assumption may be relaxed by showing $ S \succeq 0 $ with singular Hessians $ r_{ww} $ or $ r_{\theta\theta} $, through a more convoluted but relaxed requirements \citep[Section A.5.5]{Boyd2004}.

\paragraph{Proof of \cref{th:convergence}}
By \cref{prop:monotonicity} game \eqref{eq:game} is monotone.
Using \cite[Theorem 12.1.2]{facchinei2007finite} the projected gradient descent algorithm converges.

\end{document}